\documentclass[openacc]{rstransa}
\usepackage{epigraph}
\usepackage{amsmath}
\usepackage{algorithmicx}
\usepackage[ruled]{algorithm}
\usepackage{algpseudocode}
\usepackage{algpascal}
\usepackage{algc}
\usepackage{color}
\usepackage{crop}

\usepackage{amssymb,amsmath,graphicx}
\usepackage[english]{babel}
\newcommand{\Real}{\mathbb{R}}
\newcommand{\bfx}{\boldsymbol{x}}
\newcommand{\bfy}{\boldsymbol{y}}

\definecolor{redcolor}{rgb}{0.7,0.3,0.3}

\newtheorem{theore}{\bf Theorem}[section]
\newtheorem{defn}{\bf Definition}[section]
\newtheorem{proposition}{\bf Proposition}[section]

\newtheorem{corollary}{\bf Corollary}[section]

\newcommand{\Cov}{\mathrm{Cov}}
\newcommand{\Natural}{\mathbb{N}}

\renewcommand{\titlehead}{Review}

\begin{document}

\title{Blessing of dimensionality: mathematical foundations of the statistical physics of data}

\author{
A.N. Gorban$^{1}$, I.Y. Tyukin$^{2}$}

\address{$^{1}$Department of Mathematics
University of Leicester, Leicester
LE1 7RH, UK\\ $^{2}$ Department of Mathematics
University of Leicester, Leicester
LE1 7RH, UK and Department of Automation and Control Processes, Saint-Petersburg State Electrotechnical University, Saint-Petersburg, 197376, Russia}

\subject{Statistical Mechanics, Machine Learning}

\keywords{measure concentration; extreme points; ensemble equivalence;  Fisher's discriminant; linear separability}

\corres{I.Y. Tyukin\\
\email{I.Tyukin@le.ac.uk}}

\begin{abstract}
The concentration of measure phenomena were discovered as the mathematical background of statistical mecha\-nics at the end of the XIX - beginning of the XX century and were then explored in mathematics of the XX-XXI centuries. At the beginning of the XXI century, it became clear that the proper utilisation of these phenomena in machine learning might transform the {\em curse of dimensionality} into the {\em blessing of dimensionality}.

This paper summarises recently discovered pheno-mena of measure concentration which drastically simplify some machine learning problems
in high dimension, and allow us to correct legacy artificial intelligence systems. The classical concentration of measure theorems state that
i.i.d. random points are concentrated in a thin layer near a surface (a sphere or equators of a sphere, an average or median level set of energy
or another Lipschitz function, etc.).

The new {\em stochastic separation theorems} describe the thin structure of these thin layers: the random  points are not only concentrated
in a thin layer but are all linearly separable from the rest of the set, even for exponentially large random sets. The linear functionals
for separation of points can be selected in the form of the linear Fisher's discriminant.

All artificial intelligence systems make errors. Non-destructive correction requires separation of the situations (samples) with errors from the samples corresponding to correct behaviour by a simple and robust classifier. The stochastic separation theorems provide us by such
classifiers and a non-iterative (one-shot) procedure for learning.
\end{abstract}



\begin{fmtext}

\end{fmtext}
\maketitle

\section{Introduction: Five ``Foundations'', from geometry to probability, quantum mechanics, statistical physics and machine learning}
\epigraph{It's not given us to foretell \\
        How our words will echo through the ages,...}{F.I. Tyutchev, English Translation by F.Jude}

The Sixth Hilbert Problem was inspired by the ``investigations on the foundations of geometry'' \cite{Hilbert1902}, i.e. by Hilbert's work ``The
Foundations of Geometry'' \cite{Hilbert1902a}, which firmly implanted the axiomatic method not only in the field of geometry, but also in other branches of mathematics.
The Sixth Problem proclaimed expansion  of the axiomatic method beyond existent mathematical disciplines, into physics and further on.

The Sixth Problem sounds very unusual and not purely mathematical.  This
may be a reason why some great works which have been inspired by this problem have no reference to it. The most famous example is the von Neumann
book \cite{VonNeumann1955} ``Mathematical foundations of quantum mechanics''. John von Neumann was the assistant of Hilbert and they worked together
on the mathematical foundation of quantum mechanics. This work was obviously in the framework of the Sixth Problem, but this framework was not
mentioned in the book.

In 1933, Kolmogorov answered the Hilbert challenge of axiomatization of the theory of probability \cite{Kolmogorov1933}. He did not cite the
sixth problem but explicitly referred to Hilbert's ``Foundations of Geometry'' as the prototype for ``the purely mathematical development'' of
the theory. But Hilbert in his 6th Problem asked for more, for ``a rigorous and satisfactory development of the method of the mean values in mathematical
physics''. He had in mind statistical physics and ``in particular the kinetic theory of gases''. The 6th chapter of Kolmogorov's
book contains a survey of some results of the author and Khinchin about independence and the law of large numbers, and the Appendix
includes a description of the 0-1 laws in probability. These are the first steps to a rigorous basis of ``the method of mean values''.
Ten years later, in 1943, Khinchin published a book ``Mathematical foundations of statistical mechanics'' \cite{Khinchin1949}. This has brought an answer to the Sixth Problem one step closer, but again without explicit reference to Hilbert's talk.
The analogy between the titles of von Neumann
and Khinchin books is obvious.

The main idea of  statistical mechanics, in its essence, can be called the {\em blessing of dimensionality}: if a system can be presented as a
union of many weakly interacting subsystems then, in the thermodynamic limit (when the number of such subsystems tends to infinity),
the whole system can be described by relatively simple deterministic relations in the low-dimensional space of macroscopic variables.
{\em More means less} -- in  very high-dimensional spaces many differences between sets and functions become negligible (vanish)
and the laws become simpler. This point of view on statistical mechanics was developed mainly by Gibbs (1902) (ensemble equivalence) \cite{Gibbs1902} but Khinchin {made the following remark} about this work: ``although the arguments are clear from the logical standpoint, they do not pretend to any analytical  rigor'', exactly in the spirit of Hilbert's request for ``a rigorous and satisfactory development''. The devil is in the detail: how should we define the thermodynamic limit and in which sense the ensembles are equivalent? For some rigorously formulated conditions, the physical statements become exact theorems.

Khinchin considered two types of background theorems: ergodic theorems and limit theorems for high-dimensional distributions.
He claimed that the foundations of statistical mechanics should be a complete abstraction from the nature of the forces. Limit theorems
utilize very general properties of distributions in high dimension, indeed, but the expectations that ergodicity is a typical and universal
property of smooth high-dimensional multiparticle Hamiltonian systems were not met \cite{MarkusMeyer1974}.  To stress that the ergodicity
problem is nontrivial, we have to refer to the Oxtoby--Ulam theorem about metric transitivity of a generic {\em continuous}
transformation, which preserves
volume \cite{OxtobyUlam1941}. (We see that typical properties of continuous transformations differ significantly from
typical properties of smooth transformations).

Various programmes proposed for the mathematical foundation of statistical mechanics were discussed, for example,  by Dobrushin
\cite{Dobrushin1997} and Batterman \cite{Batterman1998}. Despite the impressive proof of ergodicity of some systems
(hyperbolic flows or some billiard systems, for example), the Jaynes point of view \cite{Jaynes1967} on the role of ergodicity in the foundations of statistical mechanics now became dominant; the Ergodic Hypothesis is neither necessary nor sufficient
condition for the foundation of statistical mechanics (Dobrushin \cite{Dobrushin1997} attributed this opinion to Lebowitz, while Jaynes \cite{Jaynes1967} referred to Gibbs \cite{Gibbs1902}, who, perhaps, ``did not consider ergodicity as relevant to the foundation of the subject'').

Through the efforts of many mathematicians, the limit theorems from probability theory and results about ensemble equivalence from
the foundation of statistical physics were developed far enough to become the general theory of measure concentration phenomena.
Three works were especially important for our work \cite{GianMilman2000,Gromov2003,Talagrand1995}.
The book \cite{Ledoux2005} gives an introduction into the mathematical theory of measure concentration. A simple geometric introduction into this phenomena
was given by Ball \cite{Bal1997}.

Perhaps, the simplest
manifestation of measure concentration is the concentration of the volume of the high-dimensional ball near the sphere. Let $V_n(r)$ be a volume of
the n-dimensional ball of radius $r$.  It is useful to stress that the `ball' here is not necessarily
Euclidean and means the ball of {\em any} norm.
L\'evy \cite{Levy1951}  recognised this phenomenon as a very important property of geometry of
high-dimensional spaces. He also proved that
equidistributions in the balls are asymptotically equivalent in high dimensions to the Gaussian distributions
with the same mean value of squared radius.
Gibbs {de-facto used} these properties for sublevel sets of energy to demonstrate equivalence of ensembles (microcanonical distribution on the
surface of constant energy and canonical distribution in the phase space with the same mean energy).

Maxwell used the {concentration of measure} phenomenon in the following settings. Consider a rotationally symmetric probability
distribution on the $n$-dimensional unit sphere. Then its orthogonal projection on a line will be a Gaussian distribution
with small variance $1/n$ (for large $n$ with high accuracy).
This is exactly the Maxwellian distribution for one degree of freedom in a gas (and the distribution on the unit sphere is the microcanonical
distribution of kinetic energy of gas, when the potential energy is negligibly small). Geometrically it means that if we look at the
one-dimensional projections of the unit sphere then the ``observable diameter'' will be small, of the order of $1/\sqrt{n}$.

L\'evy noticed that instead of orthogonal projections on a straight line  we can use any $\eta$-Lipschitz function
$f$ (with $\|f(x)-f(y)\|\leq \eta\|x-y\|$).
Let points $x$ be distributed on a unit $n$-dimensional sphere with rotationally symmetric probability distribution. Then the values
of $f$ will be distributed `not more widely' than a normal distribution around the mean value $\mathbf{E}_f$; for all $\varepsilon>0$
$$\mathbf{P}(|f-\mathbf{E}_f|\geq \varepsilon)\leq 2\exp \left(-\frac{n\varepsilon^2}{2c\eta^2}\right),$$
where $c$ is a constant, $c\leq 9\pi^3$. Interestingly, if we use in this inequality  the {\em median value} of $f$, $M_f$, instead of the mean, then
the estimate of the constant $c$ can be decreased: $c\leq 1$. From the statistical mechanics point of view, this L\'evy Lemma
describes the upper limit of fluctuations in gas for an arbitrary observable quantity $f$. The only condition is {the} sufficient regularity of
$f$ (Lipschitz property).

Hilbert's 6th Problem influenced this stream of research either directly (Kolmogorov and, perhaps, Khinchin among others) or indirectly, through
the directly affected works. {And it keeps to transcend this influence to other areas, including high-dimensional data analysis, data mining, and machine learning.}

On the turn of the millennium, Donoho gave a lecture about main problems of high-dimensional data analysis \cite{Donoho2000} with the impressive subtitle:
``The curses and blessings of dimensionality''.  He used the term {\em curse of
dimensionality} ``to refer to the apparent intractability of systematically searching
through a high-dimensional space, the apparent intractability of accurately approximating
a general high-dimensional function, the apparent intractability of integrating
a high-dimensional function.'' To describe the blessing of dimensionality he referred to the concentration of measure phenomenon,
``which suggest that statements about very high-dimensional settings may
be made where moderate dimensions would be too complicated.'' {Anderson} et al characterised some manifestations of this phenomenon
as ``The More, the Merrier'' \cite{AndersonEtAl2014}.

{In 1997, Kainen described the phenomenon of blessing of dimensionality, illustrated them with a number of different examples in which high dimension actually
facilitated computation, and  suggested connections with geometric phenomena in high-dimensional spaces \cite{Kainen1997}.}

The claim of Donoho's talk was similar to Hilbert's talk and he cited this talk explicitly. (``My personal research experiences,
cited above, convince me of Hilbert's position, as a
long run proposition, operating on the scale of centuries rather than decades.'') The role of  Hilbert's 6th Problem in the analysis of
the curse and blessing of dimensionality was not mentioned again.

The blessing of dimensionality and the curse of dimensionality are two sides of the same coin. For example, the typical
property of a random finite set in a high-dimensional space is: the squared distance of these points to a selected point are,
with high probability close to the average (or median) squared distance.
This property drastically simplifies the expected geometry of data (blessing) \cite{Hecht-Nielsen1994,GorbanRomBurtTyu2016}
but, at the same time, makes the similarity search in high dimensions difficult and even useless (curse) \cite{Pestov2013}.

{Extension of the 6th Hilbert Problem to data mining and machine learning is  a challenging task}.  There exist no unified general definition of machine learning. Most classical texts consider machine learning  through formalisation and analysis of a set of standardised  tasks \cite{Cucker2002, FriedmanHastieTibshirani2009, Vapnik2000}. Traditionally, these tasks are:
\begin{itemize}
\item Classification -- learning to predict a categorical attribute using values of given attributes on the basis of given examples (supervised learning);
\item Regression -- learning to predict numerical attributes using values of given attributes on the basis of given examples (supervised learning);
\item Clustering -- joining of similar objects in several clusters (unsupervised learning) \cite{Wunsch2008};
\item  Various data approximation and reduction problems: linear and nonlinear principal components \cite{GorbanKegl2008}, principal graphs \cite{GorZin2010}, independent components \cite{HyvOja2000}, etc. (clustering can be also considered as a data approximation problem \cite{Mirkin2012});
\item Probability distribution estimation.
\end{itemize}

For example, Cucker and Smale \cite{Cucker2002} considered the least square regression problem. This is the problem of the best approximation of {an} unknown function $f:X\to Y$ from a random sample of pairs $(x,y)\in X\times Y$. Selection of  ``the best'' regression function means minimization of the mean square error deviation of the observed $y$ from the value $f(x)$. They use the concentration inequalities to evaluate the probability that the approximation has a given accuracy.

It is important to mention that the Cucker--Smale approach was inspired in particular by J. von Neumann: ``We try to write
in the spirit of H. Weyl and J. von Neumann's contributions to the foundations of quantum mechanics'' \cite{Cucker2002}. The J. von Neumann  book \cite{VonNeumann1955}
was a step in the realisation of Hilbert's 6th problem programme, as we perfectly know. Therefore, the Cucker--Smale ``Mathematical foundation of learning'' is a grandchild of
the 6th problem. This is the fourth ``Foundation'' (after Kolmogorov, von Neumann, and Khinchin).  Indeed, it was an attempt  to give a rigorous development of what they ``have found to be the central ideas of learning theory''. This problem statement follows Hilbert's request  for ``rigorous and satisfactory development of the method of mean values'', but this time {the} development was done for machine learning instead of mathematical physics.

Cucker and Smale followed Gauss and proved that the least squares solution enjoys remarkable statistical properties. i.e. it provides the {\em minimum variance estimate} \cite{Cucker2002}. Nevertheless, non-quadratic functionals are employed for solution of many problems: to enhance robustness, to avoid oversensitivity to outliers, to find sparse regression with exclusion of non-necessary input variables, etc. \cite{FriedmanHastieTibshirani2009,Vapnik2000}. Even non-convex quasinorms and their tropical approximations are used efficiently to provide sparse and robust learning results \cite{GorbanMirkesZinovyev2016}.  Vapnik \cite{Vapnik2000} defined a formalised fragment of machine learning using minimisation of a {\em risk functional} that is the mathematical expectation of a general loss function.

M. Gromov \cite{Gromov2011} proposed a radically different concept of  ergosystems which function  by building their ``internal structure'' out of the ``raw structures'' in the incoming flows
of signals. The essential mechanism of egrgosystem learning is goal free and independent of any reinforcement. In {a broad sense, loosely speaking, in this concept ``structure'' = ``interesting structure''} and {learning of structure} is {goal-free} and should be considered  as a structurally interesting process.

There are {many other} approaches and algorithms in machine learning, which use some specific ideas from statistical mechanics: annealing, spin glasses, etc. (see, for example, \cite{Engel2001}) and {randomization}. {It was demonstrated recently that the assignment of random parameters  should be data-dependent to provide the efficient and universal approximation property of the randomized
learner model \cite{WangLi2017}. Various methods for evaluation of the output weights of the hidden nodes after random generation of new nodes were also tested \cite{WangLi2017}. Swarm optimization methods for learning with random re-generation of the swarm (``virtual particles'') after several epochs of learning were developed in 1990 \cite{Gorban1990}. Sequential Monte Carlo methods for learning neural networks were elaborated and tested \cite{De Freitas2001}. A comprehensive overview  of the classical algorithms and modern achievements in stochastic approaches to neural networks was performed by Scardapane and Wang  \cite{ScardapaneWang2017}}.

In our paper, we do not discuss these ideas, {instead we} focus on a deep and general similarity between high-dimensional problems in learning and statistical physics. We summarise some phenome\-na of measure concentration which drastically affect machine learning problems in high dimension.

\section{Waist concentration and random bases in machine learning}

After classical works of Fisher \cite{Fisher1936} and Rosenblatt \cite{Rosenblatt1962}, linear classifiers have been considered as inception of Data Analytics and Machine Learning (see e.g.  \cite{Vapnik2000,DudaHartStork2012,Aggarwal2015},  and references therein). The mathematical machinery powering these developments is based on the concept of linear separability.

\begin{defn} Let $\mathcal{X}$ and $\mathcal{Y}$ be subsets of $\Real^n$. Recall that a linear functional $l$ on $\Real^n$ separates $\mathcal{X}$ and $\mathcal{Y}$ if there exists a $t\in\Real$ such that
\[
l(\bfx)> t > l(\bfy) \ \forall \ \bfx\in \mathcal{X}, \ \bfy\in \mathcal{Y}.
\]
A set $\mathcal{S} \subset \mathbb{R}^n$ is {\em linearly separable} if for each $\bfx\in \mathcal{S}$ there exists a linear functional $l$ such that $l(\bfx)>l(\bfy)$ for all $y\in \mathcal{S}$, $\bfy\neq \bfx$.
\end{defn}

If $\mathcal{X}\subset \Real^n$ is a set of measurements or data samples that are labelled as ``Class 1'', and $\mathcal{Y}$ is a set of data labelled as ``Class 2'' then a functional $l$ separating $\mathcal{X}$ and $\mathcal{Y}$ is the corresponding linear classifier. The fundamental question, however, is whether such functionals exist for the given $\mathcal{X}$ and $\mathcal{Y}$, and if the answer  is ``Yes'' then how to find them?

It is well-known that if (i) $\mathcal{X}$ and $\mathcal{Y}$ are disjoint, {(ii)} the cardinality, $|\mathcal{X}\cup \mathcal{Y}|$, of $\mathcal{X}\cup \mathcal{Y}$ does not exceed $n+1$, and (iii) elements of $\mathcal{X}\cup \mathcal{Y}$ are in general position, then they are vertices of a simplex. Hence, in this setting, there {always is} a linear functional $l$  separating $\mathcal{X}$ and $\mathcal{Y}$.

Rosenblatt's $\alpha$-perceptron \cite{Rosenblatt1962} used a population of linear threshold elements with random synaptic weights ($A$-elements) as layer before an $R$-element, that is a linear threshold element which learns iteratively (authors of some papers and books called the $R$-elements ``perceptrons'' and  lose the complex structure of $\alpha$-perceptron with a layer of random $A$-elements). The randomly initiated elements of the first layer can undergo selection of the most relevant elements.

{According to Rosenblatt \cite{Rosenblatt1962}, any set of data vectors becomes linear separable after transformation by the layer of  $A$-elements, if the number of these randomly chosen elements is sufficiently large. Therefore the perceptron can solve any classification problem, where classes are defined by pointing out examples (ostensive definition).  But this ``sufficiently large'' number of random elements depends on the problem and may be large, indeed.  It can grow for a classification task proportionally to the number of the examples. The perceptron with sufficiently large number of $A$-elements can approximate binary-valued functions on finite domains with arbitrary accuracy. Recently,  the bounds on errors of these approximations are derived  \cite{Kurkova2017}. It is proven that unless the number of network units  grows faster than any polynomial of the logarithm of the size of the domain, a good approximation cannot be achieved for almost any uniformly randomly chosen function. The results are obtained by application of concentration inequalities. }

The method of random projections became popular in machine learning after the Johnson-Lindenstrauss Lemma \cite{JohnsonLindenstrauss1984}, which
states that relatively large sets of $m$ vectors in a high-dimensional Euclidean space $\mathbb{R}^d$ can be
linearly mapped into a space of much lower dimension $n$ with approximate preservation of distances.
This mapping can be constructed (with high probability) as a projection on $n$ random basis vectors with
rescaling of the projection with a factor $\sqrt{d}$ \cite{Dasgupta2003}. Repeating the projection { $O(m)$}
times and selecting the best of them, one can achieve the appropriate accuracy of the distance preservation.
The number of points $m$ can be exponentially large with $n$ ($m\leq \exp(cn)$).

Two unit random vectors in high dimension are almost orthogonal with high probability. This is a simple manifestation of the so-called
{\em waist concentration} \cite{Gromov2003}. A high-dimensional sphere is concentrated near its equator.
This is obvious: just project a sphere onto a hyperplane and use the concentration argument for a ball on the hyperplane
(with a simple trigonometric factor). This seems highly non-trivial, if we ask: near which equator? The answer is: near each equator. {This answer is obvious because of rotational symmetry but it seems to be counter-intuitive.}

We call vectors $\boldsymbol{x}$, $\boldsymbol{y}$ from  Euclidean space {\em $\mathbb{R}^n$ $\varepsilon$-orthogonal} if $|(\boldsymbol{x},\boldsymbol{y})|<\varepsilon$ ($\varepsilon>0$).
Let $\boldsymbol{x}$ and $\boldsymbol{y}$ be i.i.d. random vectors distributed uniformly (rotationally invariant) on the unit sphere in
Euclidean space $\mathbb{R}^n$. Then the distribution of their
inner product satisfies the inequality (see, for example \cite{Bal1997} or \cite{GorbTyuProSof2016} and compare to Maxwellian and L\'evy's lemma):
\[
\mathbf{P}(|(\boldsymbol{x},\boldsymbol{y})|<\varepsilon)\geq 1-2\exp\left(-\frac{1}{2}n\varepsilon^2\right).
\]

\begin{proposition}
Let $\boldsymbol{x}_1,\ldots, \boldsymbol{x}_N$ be  be i.i.d. random vectors  distributed uniformly (rotationally invariant) on the unit sphere
in Euclidean space $\mathbb{R}^n$. For
\begin{center}
\begin{equation}
\boxed{
\label{e-orthogonal Basis}
N< e^{\frac{\varepsilon^2n}{4}}\left[\ln\left(\frac{1}{1-\vartheta}\right)\right]^{\frac{1}{2}}
}
\end{equation}
\end{center}
all vectors $\boldsymbol{x}_1,\ldots, \boldsymbol{x}_N$ are pairwise $\varepsilon$-orthogonal with probability $P>1-\vartheta$. \cite{GorbTyuProSof2016}
\end{proposition}
There are two consequences of this statement: (i) in high dimension there exist exponentially many pairwise
almost orthogonal vectors in $\mathbb{R}^n$,
and (ii) $N$ random vectors are $\varepsilon$-orthogonal with high probability $P>1-\vartheta$
even for exponentially large $N$  (\ref{e-orthogonal Basis}). Existence of exponentially large $\varepsilon$-orthogonal systems
in high-dimensional spaces was discovered in 1993 by Kainen and K{\r{u}}rkov{\'a} \cite{Kurkova1993}. They introduced the notion of {\em quasiorthogonal dimension}, which was immediately utilised in the problem of random indexing of high-dimensional data \cite{Hecht-Nielsen1994}.
The fact that an exponentially large random set consists of pairwise $\varepsilon$-orthogonal vectors with high probability
was demonstrated in the work \cite{GorbTyuProSof2016} and used for analysis of data approximation problem in random bases. We
show that not only such $\varepsilon$-orthogonal sets exist, but also that they are typical in some sense.

$N$ randomly generated vectors $\boldsymbol{x}_i$ will be almost orthogonal to a given data vector $\boldsymbol{y}$ (the angle between $\boldsymbol{x}$ and $\boldsymbol{y}$ will be close to
$\pi/2$ with probability close to one). Therefore, the coefficients in the approximation of $\boldsymbol{y}$ by a linear combination of $\boldsymbol{x}_i$
could be arbitrarily large and the approximation problem will be  ill-conditioned, with high probability.
The following alternative is proven for approximation by random bases:
\begin{itemize}
\item Approximation of a high-dimensional data vector by linear combinations of randomly and independently
chosen vectors requires (with high probability) generation of exponentially
large ``bases'', if we would like to use bounded coefficients in linear combinations.
\item If arbitrarily large coefficients are allowed, then the number of
randomly generated elements that are sufficient for approximation is even less than
dimension. We have to pay for such a reduction of the number of elements by ill-conditioning of the approximation
problem.
\end{itemize}
We have to choose between a well-conditioned approximation problem in exponentially large random bases and an ill-conditional problem
in relatively small (moderate) random bases. This dichotomy is fundamental, and it is a direct consequence of the waist concentration phenomenon. In what follows, we will formally present another concentration phenomenon, stochastic separation theorems \cite{GorbanTyuRom2016,GorbTyu2017}, and outline their immediate applications in AI and neuroscience.

\section{Stochastic separation theorems and their applications in Artificial Intelligence systems}

\subsection{Stochastic separation theorems}

Existence of a linear functional that separates two finite sets $\mathcal{X},\mathcal{Y}\subset \Real^n$
 is no longer obvious when $|\mathcal{X}\cup \mathcal{Y}|\gg n$. A possible way to answer both questions could be to cast the problem as a constrained optimization problem within the framework of e.g. support vector machines \cite{Vapnik2000}. The issue with this approach is that theoretical worst-case estimates of computational complexity for determining such functions are of the order $O(|\mathcal{X}\cup \mathcal{Y}|^3)$ (for quadratic loss functions); a posteriori analysis of experiments on practical use cases, however, suggest that the complexity could be much smaller and than $O(|\mathcal{X}\cup \mathcal{Y}|^3)$ and reduce to linear or even sublinear in $|\mathcal{X}\cup \mathcal{Y}|$ \cite{Chapelle2007}.

This apparent discrepancy between the worst-case estimates and a-posteriori evaluation of computational complexities can be resolved if concentration effects are taken into account. If the dimension $n$ of the underlying topological vector space is large then random finite but exponentially large in $n$ samples are linearly separable, with high probability, {for a range of practically relevant classes of distributions}.  Moreover, we show that the corresponding separating functionals can be derived using Fisher linear discriminants \cite{Fisher1936}. Computational complexity of the latter is linear in $|\mathcal{X}\cup \mathcal{Y}|$. It can be made sub-linear too in if proper sampling is used to estimate corresponding covariance matrices. As we have shown in \cite{GorbTyu2017}, the results hold for  i.i.d. random points from equidistributions in a ball, a cube, and from distributions that are products of measures with bounded support. The conclusions are based on stochastic separation theorems for which the statements for relevant classes of distributions are provided below.

\begin{figure}
\centering
\includegraphics[width=0.4\textwidth]{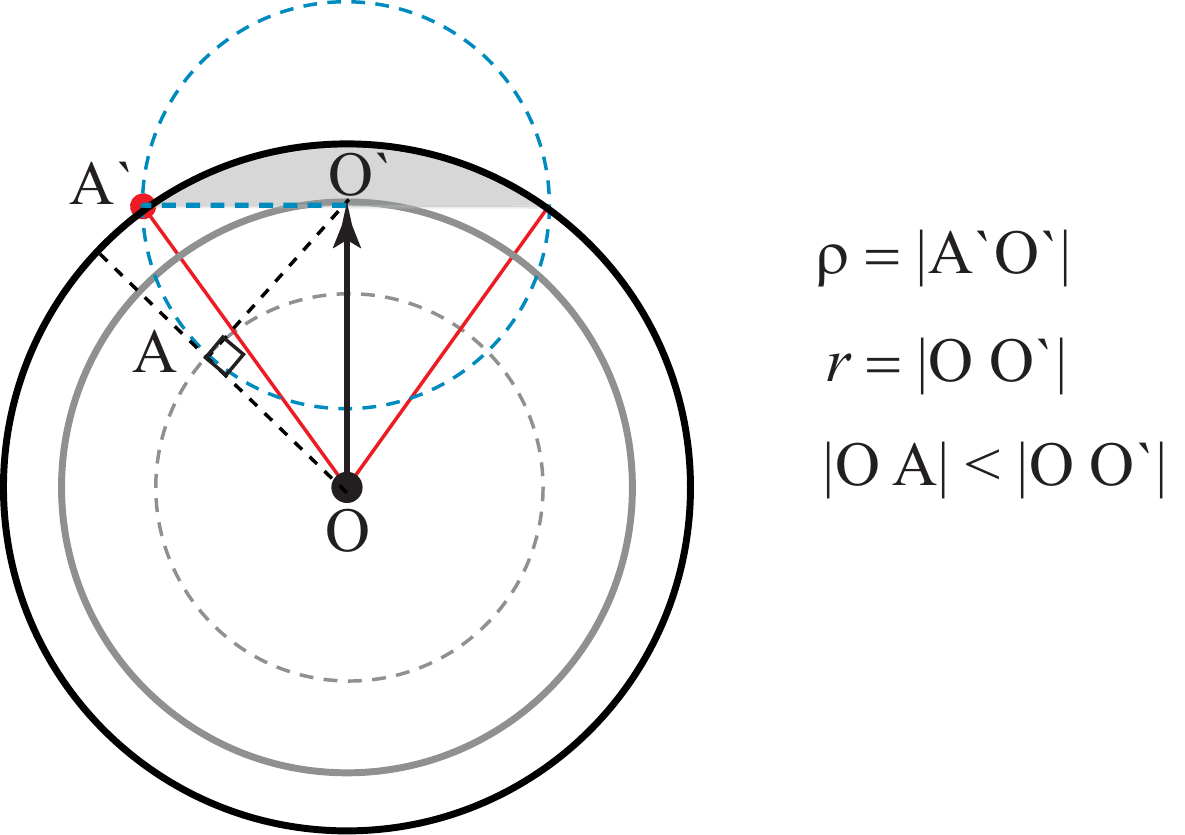}
\caption{Illustration to Theorem \ref{ball1point}.} \label{fig:theorem:ball1point}
\end{figure}

\begin{theore}[Equidistribution in $\mathbb{B}_n(1)$ \cite{GorbanTyuRom2016, GorbTyu2017}]\label{ball1point}
Let $\{\boldsymbol{x}_1, \ldots , \boldsymbol{x}_M\}$ be a set of $M$  i.i.d. random points  from the equidustribution in the unit ball $\mathbb{B}_n(1)$. Let $0<r<1$, and $\rho=\sqrt{1-r^2}$. Then
\begin{equation}\label{Eq:ball1}
\mathbf{P}\left(\|\boldsymbol{x}_M\|>r \mbox{ and } \left(\boldsymbol{x}_i,\frac{\boldsymbol{x}_M}{\| \boldsymbol{x}_M\| }\right)<r \mbox{ for all } i\neq M \right)  \geq 1-r^n-0.5(M-1) \rho^{n};
\end{equation}
\begin{equation}\label{Eq:ballM}
\mathbf{P}\left(\|\boldsymbol{x}_j\|>r  \mbox{ and } \left(\boldsymbol{x}_i,\frac{\boldsymbol{x}_j}{\| \boldsymbol{x}_j\|}\right)<r \mbox{ for all } i,j, \, i\neq j\right) \geq  1-Mr^n-0.5M(M-1)\rho^{n};
\end{equation}
\begin{equation}\label{Eq:ballMangle}
\mathbf{P}\left(\|\boldsymbol{x}_j\|>r  \mbox{ and } \left(\frac{\boldsymbol{x}_i}{\| \boldsymbol{x}_i\|},\frac{\boldsymbol{x}_j}{\| \boldsymbol{x}_j\|}\right)<r \mbox{ for all } i,j, \,i\neq j\right)  \geq  1-Mr^n-M(M-1)\rho^{n}.
\end{equation}
\end{theore}
The proof of the theorem can be illustrated with Fig. \ref{fig:theorem:ball1point}. The probability that a single element, $\bfx_{M}$, belongs to the difference $\mathbb{B}_n(1)\setminus \mathbb{B}_n(r)$ of two $n$-balls centred at $O$ is not smaller than $1-r^n$.  Consider the hyperplane
\[
l(\bfx) = r, \mbox{ where } l(\bfx)= \left(\bfx,\frac{\bfx_{M}}{\|\bfx_{M}\|}\right).
\]
This hyperplane partitions the unit ball $\mathbb{B}_n(1)$ centred at $O$ into two disjoint subsets: the spherical cap (shown as grey shaded area in Fig. \ref{fig:theorem:ball1point}) and the rest of the ball.  The element $\bfx_M$ is in the shaded area and is on the line containing the vector $OO'$. The volume of this spherical cap does not exceed the volume of the half-ball of radius $\rho$ centred at $O'$ (the ball $\mathbb{B}_n(\rho)$ is shown as a blue dashed circle in the figure). Recall that
\begin{equation}\label{eq:probability_multiple_events}
\mathbf{P}(A_1 \& A_2 \& \ldots \& A_m)\geq 1- \sum_i(1-\mathbf{P}(A_i)) \mbox{ for any events } A_1, \ldots , A_m.
\end{equation}
This assures that (\ref{Eq:ball1}) holds. Applying the same argument to all elements of the set $\mathcal{S}$ results in (\ref{Eq:ballM}). Finally, to show that (\ref{Eq:ballMangle}) holds, observe that the length of the segment $OA$ on the tangent line to the sphere $\mathbb{S}_{n-1}(\rho)$ centred at $O'$ is always smaller than $r=|OO'|$. Hence the cosine of the angle between an element from $(\mathbb{B}_n(1)\setminus \mathbb{B}_n(r)) \setminus \mathbb{B}_n(\rho)$ and the vector $OO'$ is bounded from above  by $\cos(\angle(OA',OO'))=r$. The estimate now follows from  (\ref{eq:probability_multiple_events}).

According to Theorem \ref{ball1point}, the probability that a single element $\bfx_M$ from the sample $\mathcal{S}=\{\bfx_1,\dots,\bfx_{M}\}$ is linearly separated from the set $\mathcal{S}\setminus \{\bfx_M\}$ by the hyperplane $l(x)=r$ is at least
\[
1-r^n-0.5(M-1)\left(1-r^2\right)^{\frac{n}{2}}.
\]
This probability estimate depends on both $M=|\mathcal{S}|$ and dimensionality $n$.  An interesting consequence of the theorem is that if one picks a probability value, say $1-\vartheta$, then the maximal possible values of $M$ for which the set $\mathcal{S}$ remains linearly separable with  probability that is no less than $1-\vartheta$ grows at least exponentially with $n$. In particular, the following holds

 \begin{corollary}
 Let $\{\boldsymbol{x}_1, \ldots , \boldsymbol{x}_M\}$ be a set of $M$ i.i.d. random points  from the equidustribution in the unit ball $\mathbb{B}_n(1)$. Let $0<r,\vartheta<1$, and $\rho=\sqrt{1-r^2}$. If
\begin{equation}\label{EstimateMball}
M<2({\vartheta-r^n})/{\rho^{n}},
 \end{equation}
 then
 $
 \mathbf{P}((\boldsymbol{x}_i,\boldsymbol{x}_M{)}<r\|\boldsymbol{x}_M\| \mbox{ for all } i=1,\ldots, M-1)>1-\vartheta.
$
 If
 \begin{equation}\label{EstimateM2ball}
M<({r}/{\rho})^n\left(-1+\sqrt{1+{2 \vartheta \rho^n}/{r^{2n}}}\right),
 \end{equation}
  then $\mathbf{P}((\boldsymbol{x}_i,\boldsymbol{x}_j)<r\|\boldsymbol{x}_i\| \mbox{ for all } i,j=1,\ldots, M, \, i\neq j)\geq 1-\vartheta.$

  In particular, if inequality (\ref{EstimateM2ball}) holds then the set $\{\boldsymbol{x}_1, \ldots , \boldsymbol{x}_M\}$ is linearly separable with probability $p>1-\vartheta$.
 \end{corollary}

The linear separability property of finite but exponentially large samples of random i.i.d. elements is not restricted to equidistributions in $\mathbb{B}_n(1)$. As has been noted in \cite{GorbanRomBurtTyu2016}, it holds for equidistributions in ellipsoids as well as for the Gaussian distributions. Moreover, it can be generalized to product distributions in a unit cube. Consider, e.g. {the case}  when coordinates of the vectors $\bfx=(X_1,\dots,X_n)$ in the set $\mathcal{S}$ are independent random variables $X_i$, $i=1,\dots,n$ with expectations $\overline{X}_i$ and variances $\sigma_i^2>\sigma_0^2>0$. Let $0\leq X_i\leq 1$ for all $i=1,\dots,n$. The following analogue of Theorem \ref{ball1point} can now be stated.

\begin{theore}[Product distribution in a cube \cite{GorbTyu2017}]\label{cube} Let  $\{\boldsymbol{x}_1, \ldots , \boldsymbol{x}_M\}$ be i.i.d. random points from the product distribution in a unit cube. Let
\[
R_0^2=\sum_i \sigma_i^2\geq n\sigma_0^2,
\]
and $0< \delta <2/3$. Then
\begin{equation}\label{Eq:cube1}
\begin{split}
&\mathbf{P}\left(1-\delta  \leq \frac{\|\boldsymbol{x}_j-\overline{\boldsymbol{x}}\|^2}{R^2_0}\leq 1+\delta \mbox{ and }
\left(\frac{\boldsymbol{x}_i-\overline{\boldsymbol{x}}}{R_0},\frac{\boldsymbol{x}_M-\overline{\boldsymbol{x}}}{\| \boldsymbol{x}_M-\overline{\boldsymbol{x}}\| }\right)<\sqrt{1-\delta} \mbox{ for all } i,j, \, i\neq M \right)\\
&\geq 1- 2M\exp \left(-2\delta^2 R_0^4/n \right) -(M-1)\exp \left(-2R_0^4(2-3 \delta)^2/n\right);
\end{split}
\end{equation}
\begin{equation}\label{Eq:cube2}
\begin{split}
&\mathbf{P}\left(1-\delta  \leq \frac{\|\boldsymbol{x}_j-\overline{\boldsymbol{x}}\|^2}{R^2_0}\leq 1+\delta \mbox{ and }
\left(\frac{\boldsymbol{x}_i-\overline{\boldsymbol{x}}}{R_0},\frac{\boldsymbol{x}_j-\overline{\boldsymbol{x}}}{\| \boldsymbol{x}_j-\overline{\boldsymbol{x}}\| }\right)<\sqrt{1-\delta} \mbox{ for all } i,j, \, i\neq j \right)\\
&\geq 1- 2M\exp \left(-2\delta^2 R_0^4/n \right) -M(M-1)\exp \left(-2R_0^4(2-3 \delta)^2/n\right).
\end{split}
\end{equation}
\end{theore}
The proof is based on concentration inequalities in product spaces \cite{Talagrand1995, Hoeffding1963}. Numerous generalisations of Theorems
\ref{ball1point}, \ref{cube} are possible for different classes of distributions, for example, for weakly dependent variables, etc.

Linear separability, as an inherent property of data sets in high dimension, is not necessarily confined to cases whereby a linear functional separates a single element of a set from the rest.  Theorems \ref{ball1point}, \ref{cube} be generalized to account for $m$-tuples, $m>1$ too. An example of such generalization is provided in the next theorem.

\begin{theore}[Separation of $m$-tuples  \cite{Tyukin2017a}]\label{m-tuple} Let $\mathcal{X}=\{\boldsymbol{x}_1,\dots,\boldsymbol{x}_M\}$ and $\mathcal{Y}=\{\boldsymbol{x}_{M+1},\dots,\boldsymbol{x}_{M+k}\}$  be  i.i.d. samples from the equidistribution in $\mathbb{B}_n(1)$. Let $\mathcal{Y}_c=\{\boldsymbol{x}_{M+r_1},\dots,\boldsymbol{x}_{M+r_m}\}$ be a subset of $m$ elements from $\mathcal{Y}$ such that
\begin{equation}\label{eq:k-tuples:assumption}
\beta_2 (m-1) \leq \sum_{r_j,\  r_j\neq r_i} \left( \boldsymbol{x}_{M+r_i}, \boldsymbol{x}_{M+r_j}\right)\leq \beta_1 (m-1) \ \mbox{for all} \ i=1,\dots,m.
\end{equation}
Then
\begin{equation}\label{eq:k-tuple_ball:correlated}
 \mathbf{P}\left(\exists \mbox{ a linear functional separating } \mathcal{X} \mbox{ and } \mathcal{Y}_c \right)   \geq \max_{\varepsilon\in(0,1)} (1-(1-\varepsilon)^n)^m \left(1 -
\frac{\Delta(\varepsilon,m)^\frac{n}{2}}{2}\right)^{M},
\end{equation}
where
$$ \Delta(\varepsilon,m)=1- \frac{1}{m}\left(\frac{(1-\varepsilon)^2 + \beta_2 (m-1)}{\sqrt{1+(m-1)\beta_1}}\right)^2,$$
subject to:
$$(1-\varepsilon)^2 + \beta_2 (m-1) > 0, \  1+(m-1)\beta_1 >0.$$
\end{theore}

The separating linear functional is again the inner product, and the separating hyperplane can be taken in the form \cite{Tyukin2017a}:
\begin{equation}\label{Eq:functional:m}
l(\bfx)=r; \; \mbox{ where } l(\bfx)=\left(\bfx,\frac{\bar\bfy}{\|\bar\bfy\|}\right), \; r=\frac{1}{\sqrt{m}}\left(\frac{(1-\varepsilon)^2 + \beta_2 (m-1)}{\sqrt{1+(m-1)\beta_1}}\right),
\end{equation}
and $\varepsilon$ is the maximizer of the nonlinear program in the right-hand side of (\ref{eq:k-tuple_ball:correlated}), and $\bar{\bfy}=\frac{1}{m} \sum_{i=1}^m \bfx_{M+r_i}$. To see this, observe that $\|\boldsymbol{x}_{M+r_i}\|\geq 1-\varepsilon$, $\varepsilon\in(0,1)$, for all $i=1,\dots,m$, with probability $(1-(1-\varepsilon)^n)^m$. With this probability the following estimate holds:
\[
\left(\frac{\bar{\boldsymbol{y}}}{\|\bar{\boldsymbol{y}}\|}, \boldsymbol{x}_{M+r_i} \right) \geq \frac{1}{m \|\bar{\boldsymbol{y}}\|} \left((1-\varepsilon)^2 + \beta_2 (m-1) \right).
\]
Hence
\[
\frac{1}{m}\left(1+(m-1)\beta_1 \right)\geq \left(\bar{\boldsymbol{y}},\bar{\boldsymbol{y}}\right)\geq \frac{1}{m}\left((1-\varepsilon)^2 + \beta_2 (m-1)\right),
\]
and $l(\bfx)$ in (\ref{Eq:functional:m}) is the required functional (see also Fig. \ref{fig:theorem:ball1point}).

If the elements of $\mathcal{Y}_c$ are uncorrelated, i.e. the values of $\beta_1(m-1),\beta_2(m-1)$ are  small, then the distance from the spherical cap induced by linear functional (\ref{Eq:functional:m}) to the center of the ball decreases as $O(1/\sqrt{m})$. This means that the lower-bound probability estimate in (\ref{eq:k-tuple_ball:correlated}) is expected to decrease too. On the other hand, if the elements of $\mathcal{Y}_c$ are all positively correlated, i.e. $1\geq \beta_1>\beta_2 > 0$, then one can derive  a lower-bound probability estimate which does not depend on $m$.

Peculiar properties of data in high dimension, expressed in terms of linear separability, have several consequences and applications in the realm of Artificial Intelligence and Machine Learning of which the examples are provided in the next sections.

\subsection{Correction of legacy AI systems}

Legacy AI systems, i.e. AI systems that have been deployed and are currently in operation, are becoming more and more wide-spread. Well-known commercial examples are provided by global multi-nationals, including Google, IMB, Amazon, Microsoft, and Apple. Numerous open-source legacy AIs have been created to date, together with dedicated software for their creation (e.g. Caffe \cite{Caffe}, MXNet \cite{MXNet}, Deeplearning4j \cite{Deeplearning4j}, and Tensorflow \cite{Tensorflow} packages). These AI systems require significant computational and human resources to build. Regardless of resources spent, virtually any  AI and/or machine learning-based systems are likely to make a mistake. Real-time correction of these mistakes by re-training is not always viable due to the resources involved. AI re-training is not necessarily desirable either, since AI's performance after re-training may not always be guaranteed to exceed that of the old one. {We can, therefore, formulate the technical requirements for the correction procedures. Corrector should: (i)  be simple; (ii)  not change the skills of the legacy system;  (iii) allow fast non-iterative learning; and (iv) allow correction of new mistakes without destroying of previous corrections.
}

\begin{figure}
\centering
\includegraphics[width=0.5\textwidth]{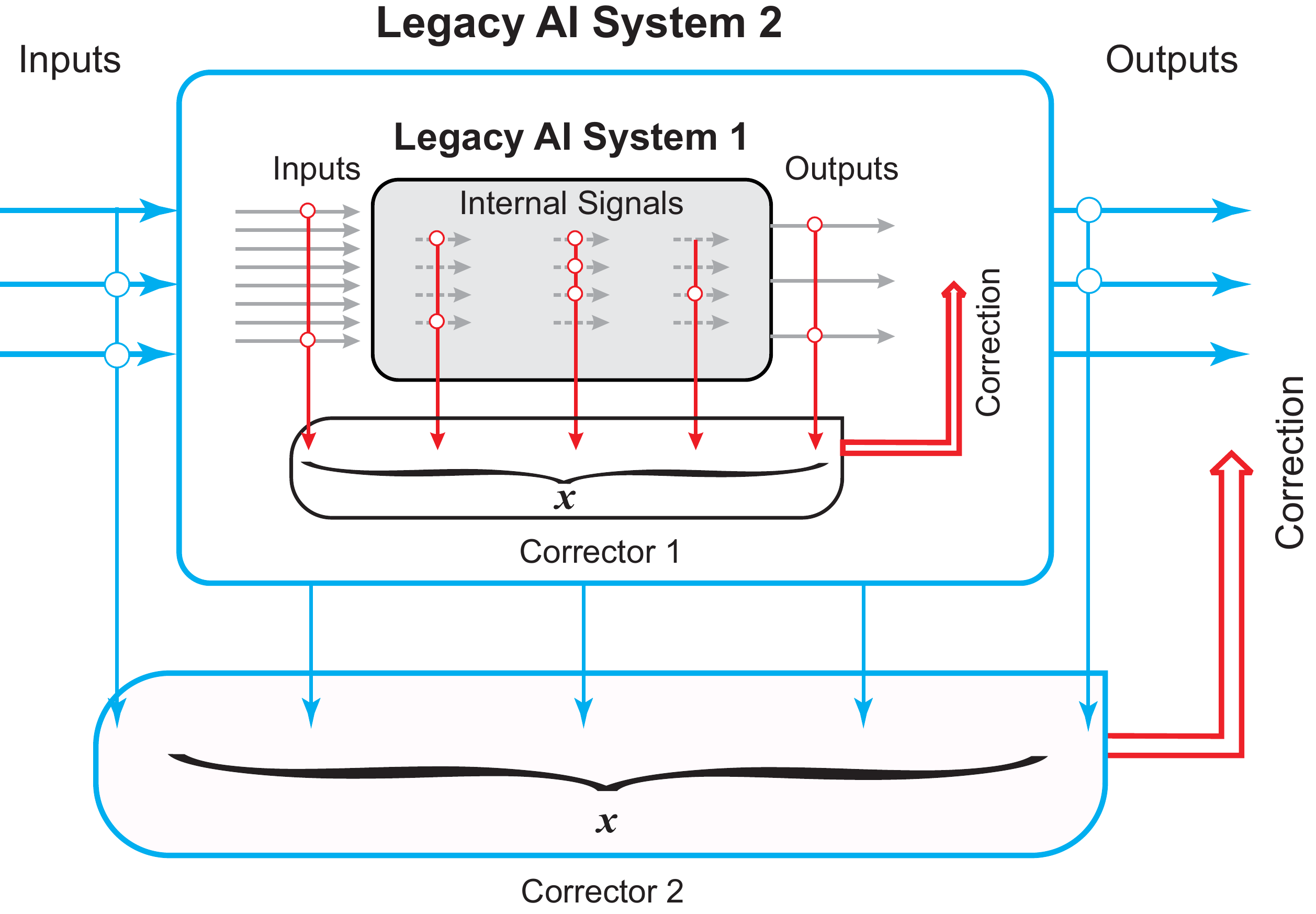}
\caption{Cascade of AI correctors}\label{fig:AI:correctors}
\end{figure}

A possible remedy to this issue is the AI correction method \cite{GorbanRomBurtTyu2016} based on stochastic  separation theorems. Suppose that at a time instance $t$ values of signals from inputs, outputs, and  internal state of a legacy AI system could be combined together to form a single measurement object, $\bfx=(x_1,\dots,x_n)$. All $n$ entries in this object are numerical values, and each measurement $\bfx$ corresponds to a relevant decision of the AI system at time $t$. Over the course of the {system's}  existence a set $\mathcal{S}$ of such measurements is collected.  For each element in the set  $\mathcal{S}$ a label ``correct'' or ``incorrect'' is assigned, depending on external evaluation of the system's performance.  Elements corresponding to ``incorrect'' labels are then filtered out and dealt with separated by an additional subsystem, a corrector. A diagram illustrating the process is shown in Fig. \ref{fig:AI:correctors}.
In this diagram, the original legacy AI system (shown as Legacy AI System 1) is supplied with a corrector altering its responses. The combined new AI system can in turn be augmented by another corrector, leading to a cascade of AI correctors (see Fig. \ref{fig:AI:correctors}).

If distributions modelling elements of the set $\mathcal{S}$ are {e.g. an equidistribution in a ball or an ellipsoid, product of measures distribution, a Gaussian etc.}, then
\begin{itemize}
 \item Theorems \ref{ball1point}--\ref{m-tuple} {guarantee} that construction of such AI correctors can be achieved using mere linear functionals.
 \item {These linear functions admit a closed-form formulae (Fisher linear discriminant) and can be determined} in a non-iterative way.
 \item {Availability of explicit closed-form formulae in the form of Fisher discriminant offers major computational benefits as it eliminates the need to employ iterative and more computationally expensive alternatives such as e.g. SVMs}.
 \item If a cascade of correctors is employed, performance of the corrected system drastically improves \cite{GorbanRomBurtTyu2016}.
\end{itemize}
{The results, perhaps, can be generalized to other classes of distributions that are regular enough to enjoy the stochastic separability property.}

\begin{figure}
\centering
\includegraphics[width=0.4\textwidth]{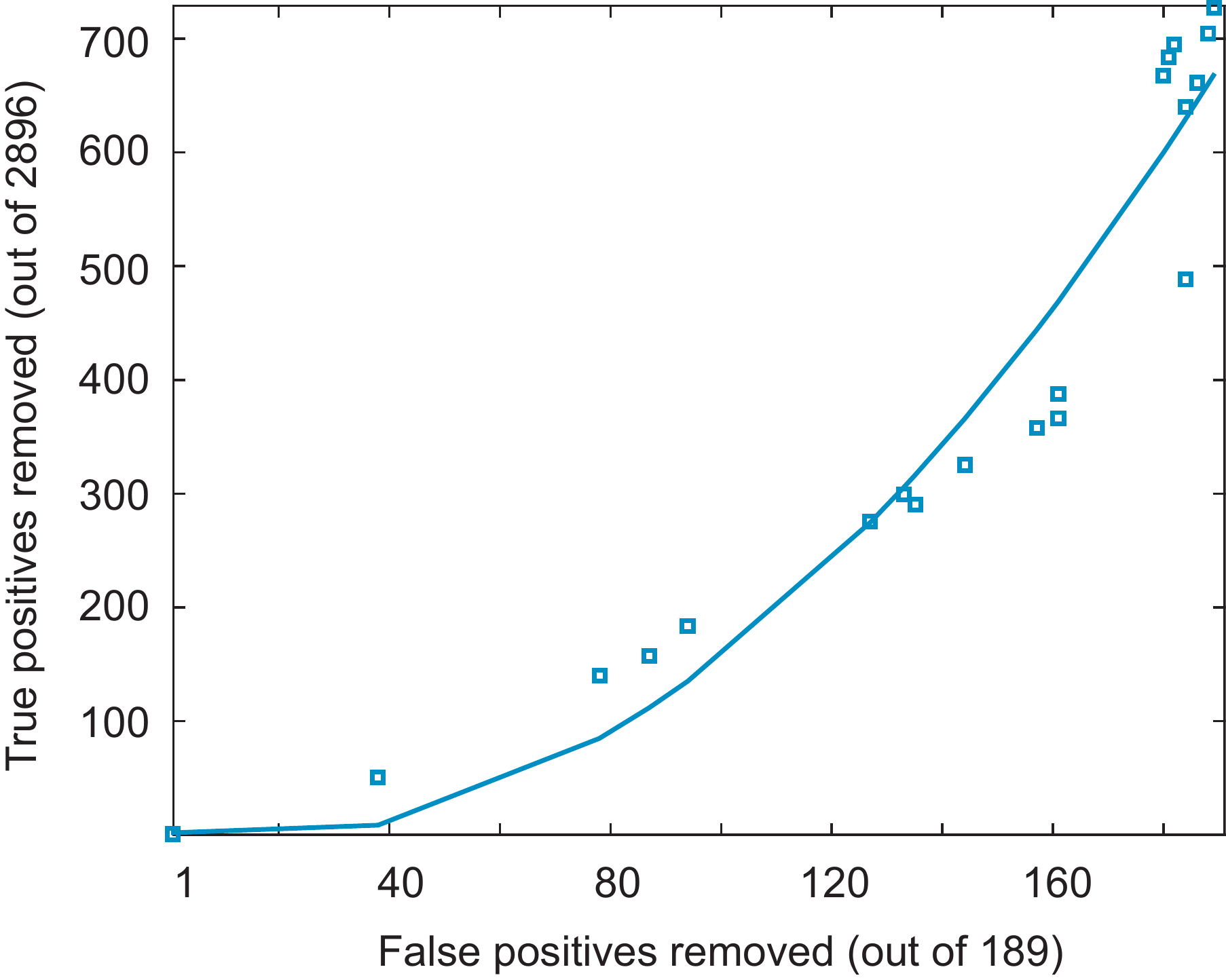}
\caption{True positives removed as a function of false positives removed by a single-functional corrector \cite{GorbanRomBurtTyu2016}.}\label{fig:corrector_experiement}
\end{figure}

The corrector principle has been demonstrated in \cite{GorbanRomBurtTyu2016} for a legacy AI system in the form of a convolutional neural network trained to detect pedestrians in images. AI errors were set to be false positives, and the corrector system had to remove labeled false positives by a single linear functional. Detailed description of the experiment is provided in \cite{GorbanRomBurtTyu2016}, and a performance snapshot is shown in Fig. \ref{fig:corrector_experiement}.
Dimensionality $n$ of the vectors $\bfx$ was $2000$. As we can see from Fig. \ref{fig:corrector_experiement}, single linear functionals are capable of removing several errors of a legacy AI without compromising the system's performance. Note that AI errors, i.e. false positives, were chosen at random and  have not been grouped or clustered to take advantage of positive correlation. (The definition of clusters could vary  \cite{Wunsch2008}.) As the number of errors to be removed grows, performance starts to deteriorate. This is in agreement with our theoretical predictions (Theorem \ref{m-tuple}).

\subsection{Knowledge transfer between AI systems}

\begin{figure}
\centering
\includegraphics[width=0.9\textwidth]{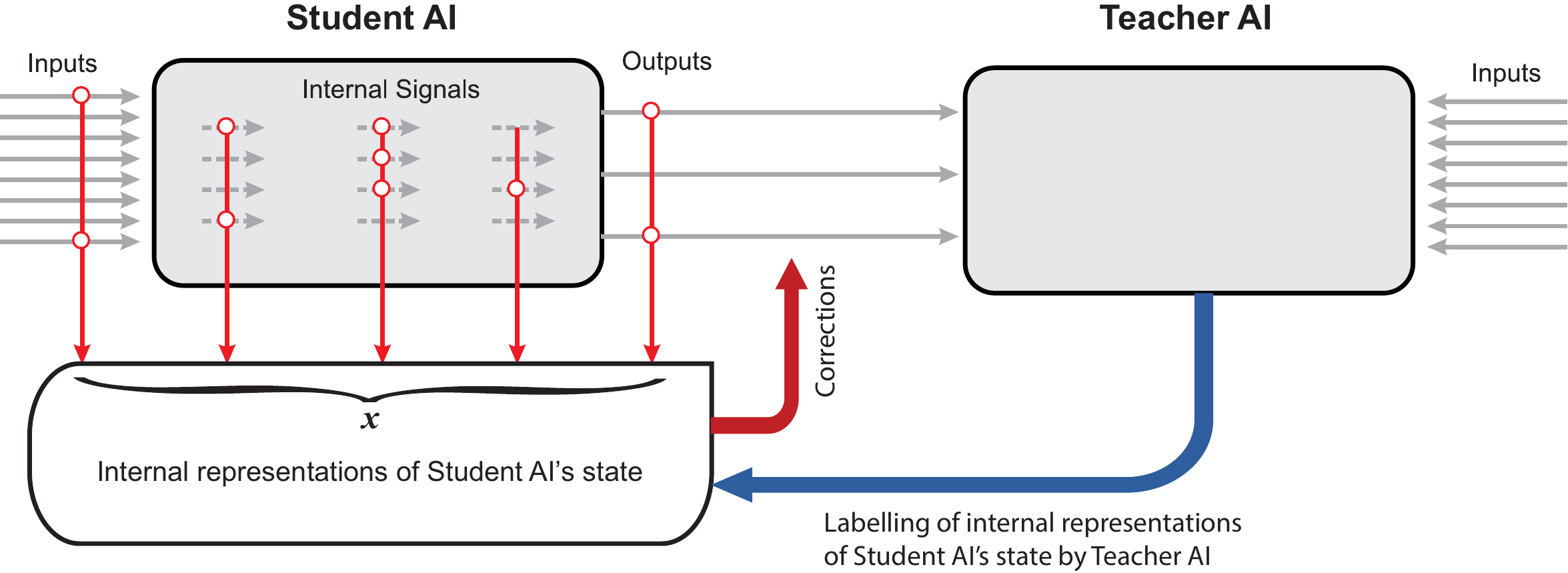}
\caption{AI Knowledge Transfer}\label{fig:AI:KT}
\end{figure}

Legacy AI correctors can be generalized to a computational framework for automated AI knowledge transfer whereby labelling of the set $\mathcal{S}$ is provided by an external AI system. AI knowledge transfer has been in the focus of growing attention during last decade \cite{Buchtala2007}. Application of stochastic separation theorems to AI knowledge transfer was proposd in \cite{Tyukin2017a}, and the corresponding functional diagram of this automated setup is shown in Fig. \ref{fig:AI:KT}.
In this setup a student AI, denoted as $\mathrm{AI}_s$, is monitored by a teacher AI, denoted as $\mathrm{AI}_t$. Over a period of activity system $\mathrm{AI}_s$ generates a set $\mathcal{S}$ of objects $\boldsymbol{x}$, $\bfx\in\Real^n$. Exact composition of the set $\mathcal{S}$ depends on a task at hand. If $\mathrm{AI}_s$ outputs differ to that of $\mathrm{AI}_t$ for the same input then an error is registered in the system. Objects $\boldsymbol{x}\in\mathcal{S}$ associated with errors are combined into the set $\mathcal{Y}$. The process gives rise to two disjoint sets:
 \[
 \mathcal{X}=\{\boldsymbol{x}_1,\dots,\boldsymbol{x}_{M}\}, \ \mathcal{X}=\mathcal{S}\setminus \mathcal{Y},
\mbox{  and  } \mathcal{Y}=\{\boldsymbol{x}_{M+1},\dots,\boldsymbol{x}_{M+k}\}.
 \]
Having created these two sets, knowledge transfer from $\mathrm{AI}_t$ to $\mathrm{AI}_s$ can now be organized in accordance with Algorithm \ref{alg:one_stage}. Note that data regularization and whitening are included in the pre-processing step of Algorithm \ref{alg:one_stage}.
\begin{algorithm}
\caption{AI Knowledge Transfer/Correction \cite{Tyukin2017a}}\label{alg:one_stage}
\small
\begin{enumerate}
  \item \textbf{Pre-processing}
  \begin{enumerate}
    \item \textit{Centering}. For the given set $\mathcal{S}$, determine the set average, $\bar{\boldsymbol{x}}(\mathcal{S})$, and generate sets $\mathcal{S}_c$
        \[
          \begin{array}{ll}
        {\mathcal{S}_c}&=\{\boldsymbol{x}\in\Real^n \ | \boldsymbol{x}=\boldsymbol{\xi}-\bar{\boldsymbol{x}}(\mathcal{S}), \ \boldsymbol{\xi}\in\mathcal{S}\}, \\
         {\mathcal{Y}_c}&=\{\boldsymbol{x}\in\Real^n \ | \boldsymbol{x}=\boldsymbol{\xi}-\bar{\boldsymbol{x}}(\mathcal{S}), \ \boldsymbol{\xi}\in\mathcal{Y}\}.
         \end{array}
        \]
    \item \textit{Regularization}. Determine covariance matrices $\Cov(\mathcal{S}_c)$, $\Cov(\mathcal{S}_c\setminus\mathcal{Y}_c)$ of the sets $\mathcal{S}_c$ and $\mathcal{S}_c\setminus\mathcal{Y}_c$. Let $\lambda_i(\Cov(\mathcal{S}_c))$, $\lambda_i(\Cov(\mathcal{S}_c\setminus\mathcal{Y}_c))$ be their corresponding eigenvalues, and $h_1, \dots, h_n$ be the eigenvectors of $\Cov(\mathcal{S}_c)$. If some of $\lambda_i(\Cov(\mathcal{S}_c))$, $\lambda_i(\Cov(\mathcal{S}_c\setminus\mathcal{Y}_c))$ are zero or if the ratio
       $ \frac{\max_i \{\lambda_i(\Sigma(\mathcal{S}_c))\}}{\min_i \{\lambda_i(\Sigma(S_c))\}}$
       is too large, project $\mathcal{S}_c$ and $\mathcal{Y}_c$  onto appropriately chosen set of $m<n$ eigenvectors, $h_{n-m+1},\dots,h_n$:
        \[
        \begin{array}{ll}
        {\mathcal{S}_r}&=\{\boldsymbol{x}\in\Real^n \ | \boldsymbol{x}=H^T \boldsymbol{\xi}, \ \boldsymbol{\xi}\in\mathcal{S}_c\}, \\
        {\mathcal{Y}_r}&=\{\boldsymbol{x}\in\Real^n \ | \boldsymbol{x}=H^T \boldsymbol{\xi}, \ \boldsymbol{\xi}\in\mathcal{Y}_c\},
        \end{array}
        \]
    where $H=\left(h_{n-m+1} \cdots h_n\right)$ is the matrix comprising of $m$ significant principal components of $\mathcal{S}_c$.
    \item \textit{Whitening}. For the centred and regularized dataset $\mathcal{S}_r$, derive its covariance matrix, $\Cov(\mathcal{S}_r)$, and generate whitened sets
  \[
  \begin{array}{ll}
  {\mathcal{S}_w}&=\{\boldsymbol{x}\in\Real^m \ | \boldsymbol{x}=\Cov(\mathcal{S}_r)^{-\frac{1}{2}} \boldsymbol{\xi}, \ \boldsymbol{\xi}\in\mathcal{S}_r\},\\
  {\mathcal{Y}_w}&=\{\boldsymbol{x}\in\Real^m \ | \boldsymbol{x}=\Cov(\mathcal{S}_r)^{-\frac{1}{2}} \boldsymbol{\xi}, \ \boldsymbol{\xi}\in\mathcal{Y}_r\}.
  \end{array}
  \]
    \end{enumerate}
   \item \textbf{Knowledge transfer}
   \begin{enumerate}
    \item \textit{Clustering}. Pick $p\geq 1$, $p\leq k$, $p\in\Natural$, and partition the set $\mathcal{Y}_w$ into $p$ clusters $\mathcal{Y}_{w,1},\dots \mathcal{Y}_{w,p}$ so that elements of these clusters are, on average, pairwise positively correlated. That is there are $\beta_{1} \geq \beta_{2} > 0$ such that:
        \[
         \beta_2(|\mathcal{Y}_{w,i}|-1)\leq \sum_{\xi\in \mathcal{Y}_{w,i}\setminus\{\boldsymbol{x}\} } \left(\boldsymbol{\xi},\boldsymbol{x}\right)  \leq \beta_1(|\mathcal{Y}_{w,i}|-1) \ \mbox{for any} \ \boldsymbol{x}\in \mathcal{Y}_{w,i}.
        \]
    \item \textit{Construction  of Auxiliary Knowledge Units}. For each cluster $\mathcal{Y}_{w,i}$, $i=1,\dots,p$, construct separating linear functionals $\ell_i$ and thresholds $c_i$:
    \[
    \begin{array}{ll}
    \ell_i(\boldsymbol{x})&=\left(\frac{\boldsymbol{w}_i}{\|\boldsymbol{w}_i\|},\boldsymbol{x}\right),\\
    \boldsymbol{w}_i&=\left(\Cov(\mathcal{S}_w \setminus \mathcal{Y}_{w,i}) + \Cov(\mathcal{Y}_{w,i}) \right)^{-1} \left(\bar{\boldsymbol{x}}(\mathcal{Y}_{w,i}) - \bar{\boldsymbol{x}}(\mathcal{S}_w\setminus\mathcal{Y}_{w,i}) \right),\\
    c_i&=\min_{\boldsymbol{\xi}\in \mathcal{Y}_{w,i}} \left( \frac{\boldsymbol{w}_i}{\|\boldsymbol{w}_i\|},\boldsymbol{\xi}\right),
    \end{array}
    \]
    where $\bar{\boldsymbol{x}}(\mathcal{Y}_{w,i})$,  $\bar{\boldsymbol{x}}(\mathcal{S}_w\setminus\mathcal{Y}_{w,i})$   are the averages of  $\mathcal{Y}_{w,i}$ and $\mathcal{S}_w \setminus \mathcal{Y}_{w,i}$, respectively. The separating hyperplane is $\ell_i(\bfx)=c_i$.

    \item \textit{Integration}. Integrate Auxiliary Knowledge Units into decision-making pathways of $\mathrm{AI}_s$. If, for an $\boldsymbol{x}$ generated by an input to $\mathrm{AI}_s$, any of $\ell_i(\boldsymbol{x})\geq c_i$ then report $\boldsymbol{x}$ accordingly (swap labels, report as an error etc.)
    \end{enumerate}
  \end{enumerate}
\end{algorithm}
The algorithm can be used for  AI correctors too. Similar to AI correction, AI knowledge transfer can be cascaded as well. Specific examples and illustrations of  AI knowledge transfer based on stochastic separation theorems are discussed in \cite{Tyukin2017a}.

\subsection{Grandmother cells, memory, and high-dimensional brain}

Stochastic separation theorems are a generic phenomenon, and their applications are not limited to AI and machine learning systems. An interesting consequence of these theorems for neuroscience has been discovered and presented in \cite{Tyukin2017}. Recently, it has been shown  that in humans new memories can be learnt very rapidly by supposedly individual neurons from a limited number of experiences \cite{Quiroga2015}. Moreover, neurons can exhibit remarkable selectivity to complex stimuli, the evidence that has led to debates around the existence of the so-called ``grandmother'' and ``concept''  cells \cite{Quiroga2005,Quiroga2009,Quiroga2012}, and their role as elements of a declarative memory. These findings suggest that not only the brain can learn rapidly but also it can respond selectively to ``rare'' individual stimuli. Moreover,  experimental evidence indicates that such a cognitive  functionality can be delivered by  single neurons \cite{Quiroga2015,Quiroga2005,Quiroga2009}. The fundamental questions, hence, are: How is this possible? and What could be the underlying functional mechanisms?

It has been shown in \cite{Tyukin2017} that stochastic separation theorems offer a simple answer to these fundamental questions. In particular, extreme neuronal selectivity and rapid learning can already be explained by these theorems. Model-wise, explanation of extreme selectivity is based on conventional and widely accepted phenomenological generic description of neural response to stimulation. Rapid acquisition of selective response to multiple stimuli by single neurons is ensured by classical Hebbian synaptic plasticity \cite{Oja1982}.

\section{Conclusion}

Twenty-three Hilbert's problems created important ``focus points'' for the concentration of efforts of mathematicians for a century.
The Sixth Problem differs significantly from the other twenty-two problems. It is very far from being a purely mathematical problem.
It seems to be impossible to imagine it's  ``final solution''. The Sixth Problem is a ``programmatic call'' \cite{Corry1997}, and it works:
\begin{itemize}
\item We definitely know that the Sixth Problem had great influence on the formulation
of the mathematical foundation of quantum mechanics \cite{VonNeumann1955}
and on the development of axiomatic quantum field theory \cite{Wightman1976}.
\item We have no doubt (but the authors have no direct evidence) that Sixth Problem has significantly affected  research in the foundation of probability theory \cite{Kolmogorov1933} and statistical mechanics \cite{Khinchin1949}.
\item The modern theory of measure concentration phenomena has direct relations to the mathematical
foundations of probability and statistical mechanics, uses results of Kolmogorov and Khinchin (among others), and
definitely helps to create  ``a rigorous and satisfactory development of the method of the mean values...''.
\item { Some of the recent attempts of rigorous approach to machine learning \cite{Cucker2002} used parts of the Sixth Problem programme \cite{VonNeumann1955}  as a prototype for their conceptual approach.}
\item The modern idea of {\em blessing of dimensionality} in high-dimensional data analysis \cite{Donoho2000, AndersonEtAl2014, GorbanTyuRom2016} is, in its essence, an extension and further development of ideas from the mathematical foundations of statistical mechanics.
\end{itemize}
The classical measure concentration theorems state that random points in a highly-dimensional data distribution
are concentrated in a thin layer near an average or median level set of a Lipschitz function.
The stochastic separation theorems describe the fine structure of these thin layers:
the random  points  are all linearly separable from the rest of the set even for exponentially large random sets.
Of course, for all these concentration and separation theorems the probability distribution should be {``genuinely''} high-dimensional.
Equidistributions in balls or ellipsoids or the products of distributions with compact support and non-vanishing variance
are the simple examples of such distributions. Various generalizations are  possible.

For which dimensions does the blessing of dimensionality work? This is a crucial question. The na\"ive point of view that
dimension of data is just a number of coordinates is wrong. This is the dimension of the dataspace, where data are originally situated.
The notion of {\em intrinsic} dimension of data is needed \cite{Kegl2003, Levina2005}. The situation when the number of data points $N$ is less (or even much less) than the dimension $d$ of the data space is not exotic. Moreover,  Donoho \cite{Donoho2000} considered the property
$d>N$ as a generic case in the ``post-classical world'' of data analysis. In such a situation we really explore
data on a $d-1$ dimensional plane and should modestly reduce our high-dimensional claim. Projection of data on that plane can be performed
by various methods. We can use as new coordinates projections of points on the known datapoints or Pearson's correlation coefficients,
when it is suitable, for example, when the datapoints are fragments of time series or large spectral images, etc. In these new coordinates
the datatable becomes a square matrix and further dimensionality reduction could be performed using good old PCA (principal component analysis), or its nonlinear versions like principal manifolds \cite{GorbanKegl2008} or neural autoencoders \cite{Bengio2009}.

A standard example can be found in \cite{Moczko2016}: the
initial dataspace consisted of fluorescence diagrams and had dimension  $5.2 \cdot 10^5$. There were 62 datapoints,
and a combination of correlation coordinates with PCA showed intrinsic dimension 4 or 5. For selection of relevant principal components
the Kaiser rule, the broken stick models or other heuristical or statistical methods can be used \cite{Cangelosi2007}.

{Similar} preprocessing ritual is {helpful} even in more ``classical'' cases {when} $d<N$. The correlation (or projection)
transformation {is not essential here}, but formation of relevant features with dimension reduction is important. If after model reduction
and {\em whitening} (transformation of coordinates to get the unit covariance matrix, { step i.c in Algorithm \ref{alg:one_stage}}) the new dimension $D\gtrsim 100$ then for $\lesssim 10^6$
datapoints we can expect that the stochastic separation theorems work with probability $>99\%$. {Thus} separation of errors with Fisher's linear discriminant is possible, {and} many other ``blessing of dimensionality benefits'' are achievable. Of course, some additional
hypotheses about the distribution functions are needed for a rigorous proof, but there is practically no
chance to check them {\em a priori} and the validation of the whole system {\em a posteriori} is necessary.
In smaller dimensions (for example, less than 10), nonlinear data approximation methods can work well {capturing} the
intrinsic  complexity of data, like principal graphs do \cite{GorZin2010, Zinovyev2013}.

We have an alternative: either essentially high-dimensional data with thin shell concentrations, stochastic separation
theorems, and efficient linear methods, or essentially low-dimensional data with efficient complex nonlinear methods.
There is a problem of the `no man's land' in-between. {To explore this land, we can extract the most interesting low-dimensional structure and then consider the residual as an essentially high-dimensional random set, which obeys stochastic separation theorems. } We do not know now a theoretically justified efficient approach to this area,
but here we should say following Hilbert: ``Wir m\"ussen wissen, wir werden wissen'' (``We must know, we shall know'').
\enlargethispage{20pt}

\competing{The authors declare that they have no competing interests.}

\aucontribute{Both authors made  substantial contributions to conception, proof of the theorems, analysis of applications, drafting the article, revising it critically, and final approval of the version to be published.}

\funding{This work was supported by Innovate UK grants KTP009890 and KTP010522. IT was supported by  the Russian Ministry of Education and Science, projects  8.2080.2017/4.6 (assessment and computational support for knowledge transfer algorithms between AI systems) and 2.6553.2017/BCH Basic Part.}


\end{document}